\documentclass{article}
\usepackage{spconf,amsmath,graphicx}
\usepackage{amsfonts}
\usepackage{multirow}
\usepackage{booktabs}
\usepackage[hidelinks]{hyperref}
\usepackage{color}
\usepackage{balance}

\title{Unravel Anomalies: An End-to-end Seasonal-Trend Decomposition Approach for Time Series Anomaly Detection}
\name{Zhenwei Zhang\textsuperscript{1}, Ruiqi Wang\textsuperscript{1}, Ran Ding\textsuperscript{2}, Yuantao Gu\textsuperscript{1}
\thanks{© 2023 IEEE. Personal use of this material is permitted. Permission from IEEE must be obtained for all other uses, in any current or future media, including reprinting/republishing this material for advertising or promotional purposes, creating new collective works, for resale or redistribution to servers or lists, or reuse of any copyrighted component of this work in other works. Code available at \url{https://github.com/zhangzw16/TADNet}}
}
\address{
\textsuperscript{1}Department of Electronic Engineering, Tsinghua University, Beijing, China \\
\textsuperscript{2}Department of Electronic Engineering, Shanghai Jiao Tong University, Shanghai, China
}

\begin{document}
\topmargin=0mm

\ninept
\maketitle

\begin{abstract}
Traditional Time-series Anomaly Detection (TAD) methods often struggle with the composite nature of complex time-series data and a diverse array of anomalies. We introduce TADNet, an end-to-end TAD model that leverages Seasonal-Trend Decomposition to link various types of anomalies to specific decomposition components, thereby simplifying the analysis of complex time-series and enhancing detection performance. Our training methodology, which includes pre-training on a synthetic dataset followed by fine-tuning, strikes a balance between effective decomposition and precise anomaly detection. Experimental validation on real-world datasets confirms TADNet's state-of-the-art performance across a diverse range of anomalies. 
\end{abstract}


\section{Introduction}
\label{sec:intro}
Time-series analysis has emerged as a pivotal area of focus across diverse application domains \cite{gu2020request, 10043819, 10.1145/3583780.3615159}. The ability to understand underlying temporal patterns and identify anomalies in these signals is paramount. Time-series Anomaly Detection (TAD) takes on the intricate task of pinpointing such deviations from expected behavior in time-series. Despite extensive efforts to accurately label data, real-world datasets are still prone to errors \cite{lai2021revisiting, SHANG2023110046}. Recognizing these challenges, our work aligns with a growing trend toward semi-supervised anomaly detection for time-series data \cite{xu2021anomaly, pang2021deep}, which operates on the assumption of a purely normal training dataset.

\textbf{Challenges.} 
TAD presents two primary challenges. First, real-world data, with its diverse origins, displays a composition of complex patterns. This necessitates models to effectively learn representations from such intricate temporal dynamics without labeled data. Second, temporal anomalies span various categories, including point- (global and contextual) and pattern-wise (shapelet, seasonal, and trend) anomalies \cite{lai2021revisiting, xu2021anomaly}. Specifically, pattern-wise anomalies can arise from varied causes, such as unusual shapes, reduced seasonality, or a declining trend. Such diversity introduces ambiguity, making accurate anomaly detection more challenging for models.

\textbf{Existing solutions.}
In recent years, deep learning models \cite{tuli2022tranad, xu2021anomaly} have surpassed classical techniques \cite{scholkopf2001estimating} in TAD tasks. These models generally fall into two categories: autoregression-based and reconstruction-based approaches. Autoregression-based methods identify anomalies through prediction errors \cite{hundman2018detecting}, while reconstruction-based methods \cite{su2019robust, li2021multivariate} use reconstruction errors. Despite their overall accuracy, many of these models fail to account for the complex compositional nature of patterns in time-series data or distinguish between different types of anomalies. Consequently, these approaches often overlook nuanced deviations and lack interpretability.

\begin{figure}[t]
\centering
\includegraphics[width=0.95\linewidth]{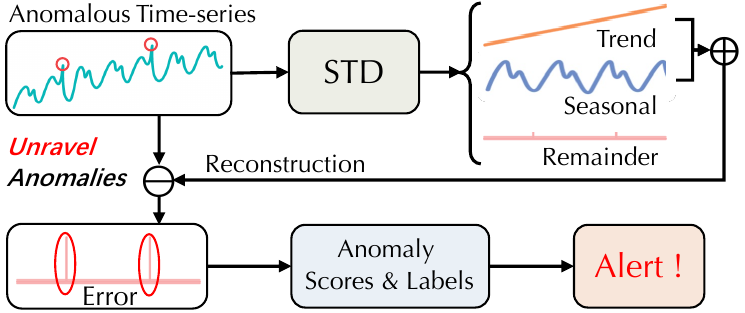}
\caption{Schematic of the STD and TAD workflow, which begins by applying STD to the anomalous time-series, yielding its decomposed components. Subsequently, the original series is compared with the reconstructed series to compute the reconstruction error. Finally, the error is transformed into anomaly scores and labels.}
\label{fig:decomp_vis}
\vspace{-15pt}
\end{figure}

\begin{figure*}[t]
\centering
\includegraphics[width=\linewidth]{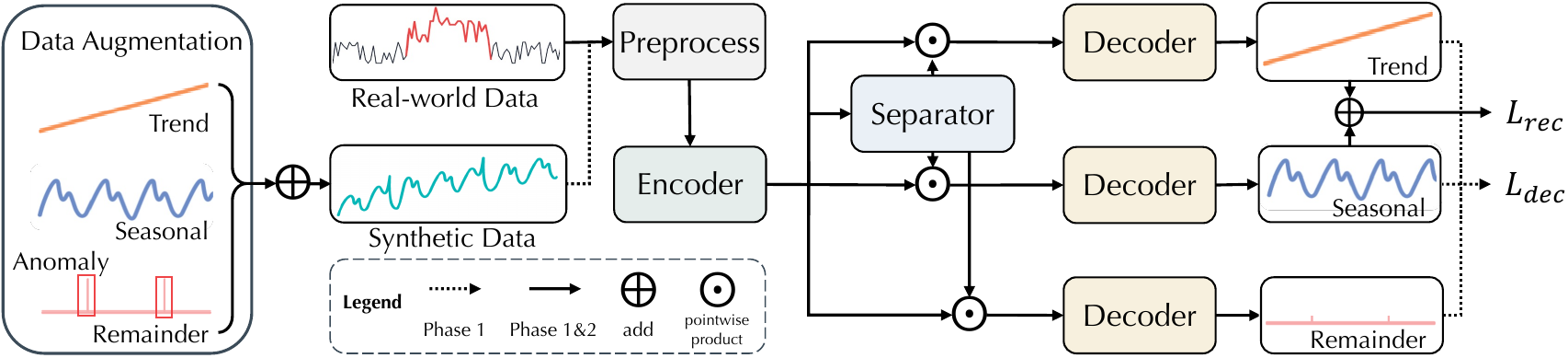}
\caption{Overview of TADNet. The workflow initiates with data augmentation, wherein trend, seasonal, anomalies, and reaminder are integrated to formulate the synthetic dataset. The training of TADnet unfolds in two phases: first, it masters STD on the synthetic data guided by \(L_{\text{dec}}\), followed by fine-tuning on real-world data, leveraging \(L_{\text{rec}}\) to capture typical patterns. The backbone of TADnet consists of an Encoder-Decoder pair, facilitating mapping between the time-domain and latent space, and a separator tasked with generating masks tailored for STD targets. The parameters are shared between the three decoders.}
\label{fig:model}
\vspace{-10pt}
\end{figure*}

\textbf{New insights.} Time series are inherently composed of multiple overlapping patterns: seasonality, trend, and remainder. This overlapping nature can obscure different types of anomalies. The advantage of using Seasonal-Trend Decomposition (STD) \cite{cleveland1990stl} is vividly demonstrated in Fig.\ref{fig:decomp_vis}, which shows how anomalies can be effectively unraveled by separating them into their respective components. By leveraging the power of STD, our approach can uniquely break down these complex composite patterns. Furthermore, following the taxonomy proposed in \cite{lai2021revisiting}, we find that different types of anomalies can be systematically associated with their respective components: seasonal anomalies with the seasonal component, trend anomalies with the trend component, and point anomalies with the remainder component.

Although existing research has incorporated time-series decomposition into TAD task \cite{qin2022decomposed, gao2020robusttad}, these approaches do not follow an end-to-end training manner. Specifically, they either depend on pre-defined decomposition algorithms, necessitating elaborate parameter tuning \cite{gao2020robusttad}, or employ decomposition only for data preprocessing \cite{qin2022decomposed}.
To overcome the lack of supervised signals for end-to-end training, we introduce a novel two-step training approach. Initially, we generate a synthetic dataset that mimics the decomposed components of real-world data. First, we pre-train our model on this synthetic dataset for decomposition tasks. The model is subsequently fine-tuned on real-world anomalous data, yielding enhanced time-series decomposition and anomaly detection.

\textbf{Contributions.} Our contributions are threefold. First, we present TADNet, a novel end-to-end TAD model that leverages STD, as illustrated in Fig.\ref{fig:model}. Inspired by TasNet \cite{luo2018tasnet}, TADNet is designed to handle time-series data similarly to audio signals, and offers detailed decomposition components, enhancing both interpretability and accuracy. Second, we adopt a targeted training strategy with pre-training and fine-tuning to enable end-to-end decomposition and detection. Third, our model achieves state-of-the-art performance and validates its efficacy through decomposition visualizations.

\section{Preliminaries}

\textbf{Time-series Anomaly Detection.} Consider a time-series \(\mathcal{T} \in \mathbb{R}^{T \times D}\) of length \(T\). The series is designated as univariate when \(D = 1\) and multivariate when \(D > 1\). The primary objective of TAD task is to identify anomalies within \(\mathcal{T}\), consequently generating an output series \(\mathcal{Y}\). Each element in \(\mathcal{Y}\) corresponds to the anomalous status of the respective data points in \(\mathcal{T}\), with $1$ indicating an anomaly. 
To facilitate this, point-wise scoring methods \cite{tuli2022tranad, schmidl2022anomaly} are employed to produce an anomaly score series \(\mathcal{S} = \{ s_1, s_2, ... , s_m \}\), each \( s_i \in \mathbb{R} \). These scores are then converted into binary anomaly labels \(\mathcal{Y}\) through an independent thresholding process.

\textbf{Seasonal-Trend Decomposition.} For a univariate time-series \(x \in \mathbb{R}^T\), its structural composition capturing trend and seasonality is represented as \(x_t = \tau_t + s_t + r_t\), where \(\tau_t, s_t\), and \(r_t\) denote the trend, seasonal, and remainder components at the \(t\)-th timestamp, respectively. 

The primary focus of our method for TAD task lies in the seasonal-trend decomposition of univariate time-series. The current literature underscores the benefits of evaluating each variable individually for increased predictive accuracy \cite{zhang2023sageformer}. Thus, each variable in a multivariate series undergoes independent decomposition, while the overall anomaly detection strategy accounts for its multivariate nature.

\textbf{Time-domain Audio Separation.} The task is formulated in terms of estimating \(C\) sources \(s^{(1)}_t, \ldots, s^{(C)}_t \in \mathbb{R}^T\), given the discrete waveform of the mixture \(x_t \in \mathbb{R}^T\). Mathematically, this is expressed as \(x_t = \sum\nolimits_{i=1}^{C} s^{(i)}_t\). 

The field of monaural audio source separation has seen advancements through various deep learning models. TasNet \cite{luo2018tasnet} introduced the concept of end-to-end learning in this area. Conv-TasNet \cite{luo2019conv} further developed this approach by incorporating convolutional layers. DPRNN \cite{luo2020dual} focused on improving long-term modeling through recurrent neural networks. More recently, architectures like SepFormer \cite{subakan2021attention} have integrated attention mechanisms.

The theoretical framework of Time-domain Audio Separation exhibits striking resemblances to the Seasonal-Trend Decomposition tasks, thus leading us to contemplate the possible transference of methodologies between these domains for improved time-series decomposition and anomaly detection.

\section{Methodology}
\label{sec:method}
\subsection{Overall Framework}

The overall flowchart of TADNet is shown in Fig.\ref{fig:model}. The preprocessing contains data normalization and segmentation. The input is first normalized to the range $[0, 1)$. Segmentation involves a sliding window approach of length \( P \), converting the normalized \( \mathcal{T} \) into non-overlap blocks of length $P$ denoted as \( \mathcal{D}=\left\{\mathcal{X}_1, \mathcal{X}_2, \ldots, \mathcal{X}_N\right\} \). Notably, while segmentation offers a more flexible approach to managing longer sequences, it does not influence results.

Within the TADNet backbone, we leverage the TasNet architecture and its variants \cite{luo2018tasnet, luo2019conv, luo2020dual} from speech separation. Viewing the seasonal and trend components as distinct audio signals, TasNet facilitates effective STD (see Fig.\ref{fig:model}).

Since the training utilizes only normal samples, anomalies typically disrupt the reconstruction process. To detect these anomalies, we compute the reconstruction error, denoted by \(Score(t) = \lVert \mathcal{T}_{t,:} - \hat{\mathcal{T}}_{t,:} \rVert_2\), where \(\lVert \cdot \rVert_2\) represents the L2 norm. 

\subsection{TADNet Backbone}
\begin{table*}[t!]
    \caption{Quantitative results for TADNet across five real-world datasets use metrics \emph{P}, \emph{R}, and \emph{F1} for precision, recall, and F1-score (\%). Higher values indicate better performance. Best and second-best results are in bold and underlined, respectively. Dataset are followed by brackets, where \emph{u} indicates univariate and \emph{m} multivariate.}\label{tab:main_results}
    \centering
    \resizebox{\textwidth}{!}{
    \renewcommand{\multirowsetup}{\centering}
    \begin{tabular}{c|ccc|ccc|ccc|ccc|ccc}
    \toprule
    Dataset & \multicolumn{3}{c|}{\textbf{UCR} (\emph{u})}& \multicolumn{3}{c|}{\textbf{SMD} (\emph{m})}& \multicolumn{3}{c|}{\textbf{SWaT} (\emph{m})}& \multicolumn{3}{c|}{\textbf{PSM} (\emph{m})} & \multicolumn{3}{c}{\textbf{WADI} (\emph{m})}\\
    Metric& P& R& F1& P& R& F1& P& R& F1& P& R& F1& P& R& F1 \\
    \midrule

    OCSVM        & 41.14 & 94.00  & 57.23 & 44.34 & 76.72 & 56.19 & 45.39 & 49.22 & 47.23 & 62.75 & 80.89 & 70.67 & 61.89 & 62.31 & 62.10 \\
    OmniAnomaly  & 64.21 & 86.93  & 73.86 & 83.34 & 94.49 & 88.57 & 86.33 & 76.94 & 81.36 & 91.61 & 71.36 & 80.23 & 31.58 & 65.41 & 42.60 \\
    InterFusion  & 60.74 & 95.20  & 74.16 & 87.02 & 85.43 & 86.22 & 80.59 & 85.58 & 83.01 & 83.61 & 83.45 & 83.52 & 80.26 & 30.38 & 44.08 \\
    AnomalyTran  & 72.80 & 99.60  & 84.12 & 89.40 & 95.45 & \underline{92.33} & 91.55 & 96.73 & \textbf{94.07} & 96.91 & 98.90 & \underline{97.89} & 80.30 & 79.23 & 79.76 \\
    TranAD       & 94.07 & 100.00 & \underline{96.94} & 88.03 & 89.42 & 88.72 & 97.60 & 69.97 & 81.51 & 96.44 & 87.37 & 91.68 & 35.29 & 82.96 & 49.51 \\
    DecompTran   & 71.58 & 96.83  & 82.31 & 89.32 & 93.94 & 91.57 & 95.17 & 80.30 & 87.10 & 97.65 & 87.21 & 92.14 & 79.40 & 81.01 & \underline{80.20} \\

    \midrule
    \textbf{TADNet(Ours)}& 97.51 & 100.00 & \textbf{98.74} & 94.81 & 91.93 & \textbf{93.35} & 92.15 & 88.35 & \underline{90.21} & 98.12 & 99.21 & \textbf{98.66} & 94.03 & 82.96 & \textbf{88.15} \\
    \bottomrule
    \end{tabular}
    }
    \vspace{-10pt}
\end{table*}

The encoder accepts a univariate time series \(x_d\in\mathbb{R}^{P}\), where \(d=1,2,\ldots,D\), sourced from multivariate \(\mathcal{X}_i\). It segments this series into multiple overlapping frames. Each frame possesses a length of \( L \) and overlaps with adjacent frames by a stride of \( S \). Sequentially, these frames are collated to constitute \( \mathbf{X}_d \in \mathbb{R}^{L \times K} \). Through a subsequent linear transformation, the encoder projects \( \mathbf{X}_d \) into a latent space, given by:
$\mathbf{E} = \mathbf{U}\mathbf{X}_d.$
The matrix \( \mathbf{U} \in \mathbb{R}^{N \times L} \) contains trainable transformation bases in its rows, while \( \mathbf{E} \in \mathbb{R}^{N \times K} \) represents the feature representation of the input time series in the latent space.


The separator receives the encoded representation and is tasked with generating masks for each of the decomposed components. Formally, \(\{ \mathbf{M}_{\tau}, \mathbf{M}_s, \mathbf{M}_r \} = \mathcal{F}_{\text{sep}}(\mathbf{E}; \theta)\), where \(\mathbf{M}_{\tau}\), \(\mathbf{M}_s\), and \(\mathbf{M}_r\) represent the masks for the trend, seasonal, and remainder components, respectively. Here, \(\mathcal{F}_{\text{sep}}\) denotes the separator subnetwork, which can be implemented using various architectures such as CNN \cite{luo2019conv}, RNN \cite{luo2020dual}, or Transformer \cite{subakan2021attention}.
Utilizing these masks, the embeddings for each target from the global feature \( \mathbf{E} \) are:
\begin{equation}
\mathbf{E}_{\tau}= \mathbf{M}_{\tau} \odot \mathbf{E}, \quad
\mathbf{E}_s= \mathbf{M}_s \odot \mathbf{E}, \quad
\mathbf{E}_r= \mathbf{M}_r \odot \mathbf{E},
\end{equation}
achieved via point-wise products of the corresponding masks.

The decoder architecture mirrors the encoder, taking in the masked embeddings generated by the separator. These embeddings are mapped back to the time domain via a linear transformation \( V \):
\begin{equation}
\hat{\mathbf{S}}_{\tau} = \mathbf{E}_{\tau}^T \mathbf{V},\quad
\hat{\mathbf{S}}_s = \mathbf{E}_s^T \mathbf{V},\quad
\hat{\mathbf{S}}_r = \mathbf{E}_r^T \mathbf{V}.
\end{equation}
Here, \( V\in\mathbb{R}^{N\times L}\) has \( N \) decoder bases. The reconstructed trend, seasonal, and remainder, denoted as \( \hat{\mathbf{S}}_{\tau} \), \( \hat{\mathbf{S}}_s \), and \( \hat{\mathbf{S}}_r \), are derived from their respective embeddings. The output time-domain signals, $\hat\tau_d$, $\hat s_d$, and $\hat r_d$, are obtained through an overlap-and-add operation.

\subsection{Synthetic Dataset}

In anomaly detection, real-world data often lack the nuanced trend and seasonal patterns essential for STD. To equip TADnet with robust STD capabilities, we constructed a composite dataset. This dataset is meticulously designed with intricate seasonal and trend shifts, anomalies, and noise to emulate real-world contexts, as illustrated in Fig.\ref{fig:decomp_vis}. Both deterministic and stochastic trends are leveraged to craft the trend and seasonal components, which are subsequently normalized to maintain a zero mean and unit variance.

\textbf{Trend.}
The deterministic trend is generated using a linear trend function with fixed coefficients: \( \tau_t^{(d)} = \beta_0 + \beta_1 \cdot t \), where \( \beta_0 \) and \( \beta_1 \) are tunable parameters. The stochastic trend component is modeled using an ARIMA(0,2,0) process, integrated into the trend model as follows:  $\tau_t^{(s)}=\sum_{n=1}^t n X_n$, where \( X_t \) is a normally-distributed white noise term, satisfying \( \Delta^2 \tau^{(s)}_t = X_t \).

\textbf{Seasonal.}
The deterministic seasonal component combines various types of periodic signals. It includes sinusoidal waves with varying amplitudes, frequencies, and phases, as well as square waves with different amplitudes, periods, and phases. 

For the slow-changing stochastic sequence, the seasonal component is composed of repeating cycles of a slow-changing trend series \( \tau^{(s)}_t \). This series is generated using the trend generation algorithm to ensure a smooth transition between cycles. Each cycle is uniquely characterized by a period \( T_0 \) and a phase \( \phi \). The stochastic seasonal component is thus formulated as \( s_t = \tau^{(s)}_{\text{mod}(t + \phi, T_0)} \).

To enrich the dataset, minor adjustments are made to both the cycle's length and amplitude, including resampling individual cycles and scaling values within a cycle, aiming for more diverse and generalizable signal decomposition.

\textbf{Remainder.}
The remainder component is conceived using a white noise process with adjustable variances.

To enhance the robustness of the decomposition model against anomalies and ensure stable decomposition performance, we injected a portion of anomalous data into the synthetic dataset, following the method outlined in \cite{lai2021revisiting}.

\subsection{Two-Phase Training Strategy}
We propose a two-phase training strategy for TADnet to ensure its efficacy in both time-series decomposition and anomaly detection tasks.

In the first phase, TADnet is pre-trained on a synthetic dataset, with a focus on time-series decomposition. The corresponding loss function, which aggregates the Mean Squared Errors for each decomposed component, is formulated by:
\begin{equation}
    L_{\text{dec}} = \sum\nolimits_{d=1}^{D} \left( \lVert \tau_d-\hat\tau_d \rVert_2^2 + \lVert s_d-\hat s_d \rVert_2^2 + \lVert r_d-\hat r_d \rVert_2^2 \right)
\end{equation}
Here, \( \tau_d \), \( s_d \), and \( r_d \) denote the actual seasonal, trend, and residual components for the \( d \)-th dimension, respectively, while \( \hat{\tau}_d \), \( \hat{s}_d \), and \( \hat{r}_d \) represent their predicted counterparts.

In the second phase, TADnet is fine-tuned using a real-world TAD dataset. This stage emphasizes the accurate reconstruction of the original time series following its decomposition, a key requirement for effective anomaly detection. The loss function for this stage, which focuses on overall reconstruction accuracy, is given by:
\begin{equation}
L_{\text{rec}} = \sum\nolimits_{d=1}^{D} \lVert x_d- (\hat\tau_d + \hat s_d)\rVert_2^2
\end{equation}
Here, \( x_d \) represents the original time series in the \( d \)-th dimension, and \( \hat{\tau}_d + \hat{s}_d \) is its predicted reconstruction.

\begin{table}[thbp]
\vspace{-10pt}
\caption{Details of the dataset.}\label{tab:dataset}
    \centering
    \resizebox{\columnwidth}{!}{
    \begin{tabular}{c|rr|r|r|r}
      \toprule
      Dataset & \#Entities & \#Dim & \#Train & \#Test(labeled) & Anomaly\%  \\
      \midrule
       UCR        & 4  & 1  & 1,200-3,000    & 4,500-6,301    & 1.9\\
       SMD        & 28 & 38 & 23.6K-28.7K & 23.6K-28.7K & 4.2 \\
       WADI       & 1  & 127 & 789,371 & 172,801 & 5.9\\
       SWaT       & 1  & 51 & 496,800 & 449,919 & 12.1\\
       PSM        & 1  & 25 & 132,481 & 87,841  & 27.8\\
      \bottomrule
    \end{tabular}}
\vspace{-10pt}
\end{table}

\section{Experiments}

\begin{figure*}[t]
\centering
\includegraphics[width=0.96\linewidth]{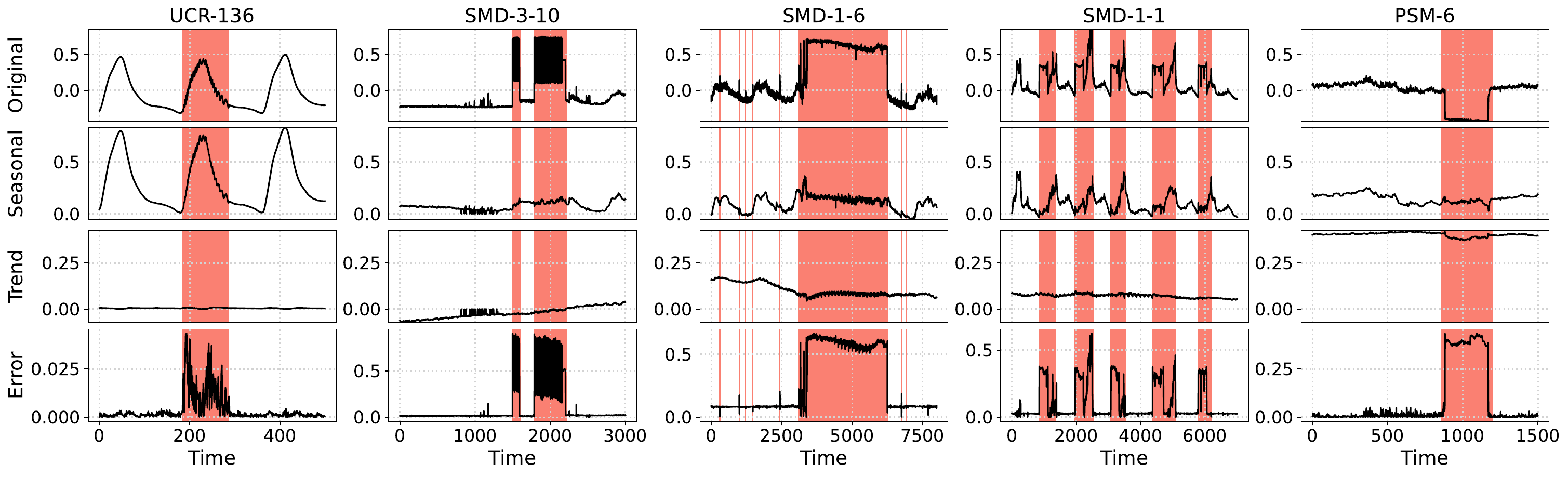}
\vspace{-10pt}
\caption{Visualization of decomposition and detection results in UCR and SMD. The first row shows the raw time series with anomalies, the second and third rows display the seasonal and trend components, respectively, and the final row depicts the reconstruction error. Anomalies are marked with a red background.}
\label{fig:real_results}
\end{figure*}

In our experiments, we employ five real-world datasets encompassing both univariate and multivariate time-series, as outlined in Table \ref{tab:dataset}. 
These include multi-univariate dataset UCR \cite{keogh2021multi}, featured in the KDD 2021 Cup; SMD \cite{su2019robust}, which provides five weeks of data from a leading Internet company; and SWaT \cite{mathur2016swat}, offering sensor data from water treatment plant. WADI extends SWaT but contains over twice as many sensors and actuators. Additionally, PSM \cite{abdulaal2021practical} originates from eBay's application server nodes.
NASA's MSL and SMAP \cite{hundman2018detecting} datasets were excluded due to their abundance of binary sequences, which are incompatible with our decomposition methods.

We benchmark TADNet against: one classic method, OCSVM \cite{scholkopf2001estimating}, and five deep models including OmniAnomaly \cite{su2019robust}, InterFusion \cite{li2021multivariate}, AnomalyTran \cite{xu2021anomaly}, TranAD \cite{tuli2022tranad}, and DecompTran \cite{qin2022decomposed}. Other traditional methods are excluded, as deep learning models have been proven superior \cite{xu2021anomaly}. Evaluation is based on standard TAD metrics such as precision, recall, and F1 score. We adopt a widely-used adjustment strategy \cite{su2019robust, xu2021anomaly, tuli2022tranad, qin2022decomposed}: if any time point in an abnormal segment is detected, the entire segment is considered correctly identified, aligning with real-world applications.

We partition the dataset into blocks of \(P=8,000\) time points to optimize memory. The backbone utilizes TasNet with DPRNN \cite{luo2020dual}, featuring kernel size \(W=2\), encoding dimension \(E=256\), feature dimension \(F=64\), hidden dimension \(H=128\), and six layers. Training employs ADAM with an initial learning rate of \(1 \times 10^{-3}\) for 200 epochs, followed by fine-tuning at \(5 \times 10^{-4}\) for 10-20 epochs. Experiments run on an NVIDIA GeForce RTX 3090. To ensure fair comparisons with previous work, the Peak Over Threshold method \cite{siffer2017anomaly} is employed to determine the anomaly threshold. A timestamp is labeled as anomalous if its score exceeds this threshold.

\subsection{Results and Analysis}
Our study presents the experimental outcomes in Table \ref{tab:main_results}, where TADNet is rigorously evaluated against six baselines on five real-world datasets. Demonstrating outstanding performance, TADNet notably achieves the highest F1-score in four of the datasets. The application of seasonal trend decomposition in our method effectively discerns complex temporal patterns, broadening the anomaly detection scope. These results empirically substantiate the effectiveness of decomposition techniques in time-series anomaly detection. Additionally, while TADNet's two-stage pre-training extends the training duration to less than an hour, its inference time aligns with that of benchmark algorithms.

\textbf{Visualization.}
To demonstrate the effectiveness of STD in unraveling complex anomalies, we showcase decomposition and reconstruction error results on multiple real-world datasets in Fig.\ref{fig:decomp_vis} (NeurIPS-TS) and Fig.\ref{fig:real_results} (UCR and SMD). It's noteworthy that anomalies become more clearly evident when comparing their respective decomposed components to the original series. As the reconstruction error and anomaly scores are positively correlated, our reconstruction error closely aligns with the anomalous regions, as seen in the last row. This validates that our method adeptly identifies anomalies, thereby improving detection accuracy while reducing false positives.

\begin{table}[t]
\vspace{-10pt}
\caption{Ablation results, evaluated using F1-score (\%). '\textit{w/o Sep}' excludes the separator from the backbone architecture; '\textit{w/o Decomp}' replaces \(L_{\text{dec}}\) with \(L_{\text{rec}}\); '\textit{w/o Augment}' omits pretraining on a synthetic dataset; and '\textit{Iterative}' involves iterative training between the decomposition and anomaly detection tasks.}\label{tab:ablation}
\centering
\resizebox{0.9\linewidth}{!}{
\renewcommand{\multirowsetup}{\centering}
\begin{tabular}{c|ccccc}
\toprule
Ablation & UCR & SMD & SWaT & PSM & WADI \\
\midrule
TADNet              & 98.74          & \textbf{93.35} & \textbf{90.21} & \textbf{98.66} & 88.15          \\
\midrule
\textit{w/o Sep}    & 32.68          & 66.24          & 76.89          & 83.28          & 47.66          \\
\textit{w/o Decomp} & 48.69          & 84.12          & 88.41          & 95.57          & 65.72          \\
\textit{w/o Augment}& 40.12          & 74.17          & 83.26          & 98.01          & 62.15          \\
\textit{Iterative}  & \textbf{99.12} & 92.14          & 86.55          & 96.58          & \textbf{92.06} \\
\bottomrule
\end{tabular}
}
\vspace{-10pt}
\end{table}

\textbf{Ablation Study.}
As shown in Table \ref{tab:ablation}, we further investigate the effect of each part in TADNet. The removal of key components such as the Separator (\textit{w/o Sep}), Decomposition (\textit{w/o Decomp}), or Augmentation (\textit{w/o Augment}) leads to substantial drops in F1-scores, highlighting their essential roles in TADNet's performance. While the iterative training approach (\textit{Iterative}) shows some improvement in specific datasets (notably WADI, increased to 92.06\%), its computational overhead makes it less practical for broader applications. We therefore opt for the pretrain-finetune paradigm and leave the study of iterative training for future work.

\section{Conclusion}
In this work, we present TADNet, an end-to-end TAD model that employs STD to tackle the challenges of complex patterns and diverse anomalies. Our model distinguishes itself by providing interpretable and accurate decomposition components. The model's effectiveness is enhanced through a dual-phase training approach, initially using synthetic data and subsequently fine-tuning with real-world data, achieving top-tier performance with clear decomposition visualizations. While primarily effective for univariate time series, TADNet extends to multivariate series, though its accuracy improvement in certain complex multivariate scenarios may be limited.
%



\vfill\pagebreak
\bibliographystyle{IEEEbib}
{
\ninept
\bibliography{refs}
}

\end{document}